# Revisiting the Efficacy of Signal Decomposition in AI-based Time Series Prediction

Kexin Jiang[a], Chuhan Wu[b], Yaoran Chen[c,*]

a Department of Computer Engineering and Science, Shanghai University, 99 Shangda Road, Baoshan District, Shanghai, 200444, China
b Department of Electronic Engineering, Tsinghua University, 30 Shuangqing Road, Haidian District, Beijing, 100084, China
c Institute of Artificial Intelligence, Shanghai University, 99 Shangda Road, Baoshan District, Shanghai, 200444, China

**Abstract**

Time series prediction is a fundamental problem in scientific exploration and artificial intelligence (AI) technologies have substantially bolstered its efficiency and accuracy. A well-established paradigm in AI-driven time series prediction is injecting physical knowledge into neural networks through signal decomposition methods, and sustaining progress in numerous scenarios has been reported. However, we uncover non-negligible evidence that challenges the effectiveness of signal decomposition in AI-based time series prediction. We confirm that improper dataset processing with subtle future label leakage is unfortunately widely adopted, possibly yielding abnormally superior but misleading results. By processing data in a strictly causal way without any future information, the effectiveness of additional decomposed signals diminishes. Our work probably identifies an ingrained and universal error in time series modeling, and the de facto progress in relevant areas is expected to be revisited and calibrated to prevent future scientific detours and minimize practical losses.

∗Corresponding author

ychen169@shu.edu.cn (Yaoran Chen)

# 1. Introduction

Time series, a fundamental data form composed of successive sequences of data points, is critical in many time-sensitive areas such as meteorology [1], economics [2], environics [3] and sociology [4]. Predicting the future trends of time series based on past observations enables researchers to understand the future patterns of natural and societal signals at different time scales, ranging from long-term sunspot activities and flares [5], hour-level tropical cyclone trajectories [6], and near-real-time fluctuations of volatile renewable energy like solar [7] and wind [8]. Thus, accurate time series prediction has tremendous potential for various research and industrial applications, such as decision-making [9], resource allocation [10], business planning [11], and risk management [12].

In the realm of physical-related areas such as oceanography and meteorology, canonical time series prediction models rely on numerical methods derived from physical modeling [13, 14]. Although these methods have been extensively used and improved over the decades [15, 16], the heavy computational costs hinder their applications in large-scale and real-time systems, and contradict the trend of green computing. To address this challenge, artificial intelligence (AI)-based methods have emerged as an alternative approach for predicting natural time series, offering the advantages of high efficiency, flexibility, and scalability [17]. With the rise of deep learning, natural time series prediction models based on deep neural networks have garnered widespread attention and achieved impressive results [18]. This technology trend has impacted a broader interdisciplinary community that involves data-driven time series analytics [19].

Deep learning time series models have demonstrated their effectiveness in discovering rich and complex patterns hidden in big data [20, 21]. However, these models are often criticized for their lack of prior knowledge and awareness of physical laws, resulting in non-robust predictions [22, 23]. To handle this issue, researchers have proposed to explicitly inject manually crafted information into AI models to empower them with accumulated domain knowledge [24, 25]. A representative example is time-frequency decomposition, which can help AI models deal with highly volatile and noisy time series data [26, 27]. Through learning on stabilized decomposition signals, models can better capture inherent patterns that match physical phenomena [28, 29]. Consequently, a considerable amount of work along this research line has been extensively conducted in various areas [30], with encouraging and even striking results reported in many downstream applications [31, 32, 33].

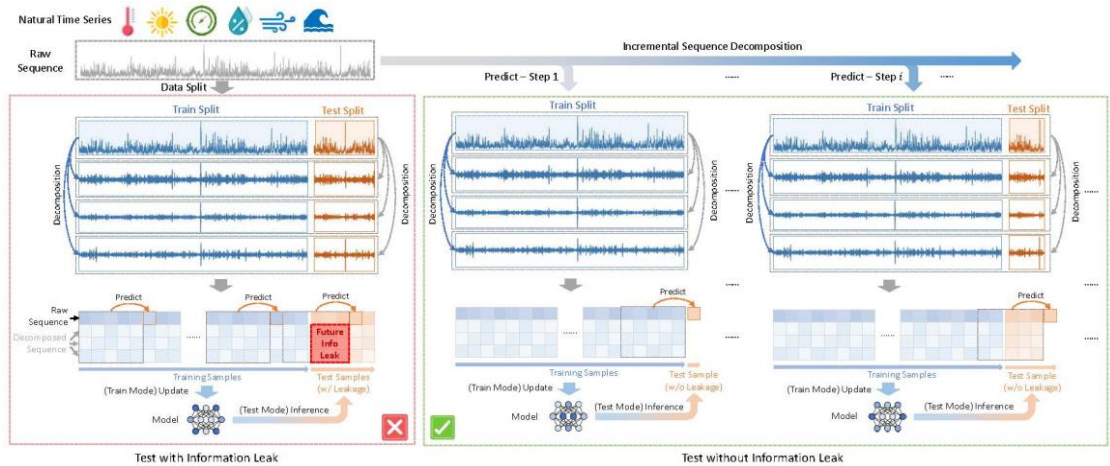

Figure 1: The data leakage problem of leveraging signal decomposition in time series prediction. A widely employed method is performing sequence decomposition on the entire sequence, and then dividing the decomposed sequences into training and test splits. When training on the decomposed sequences, the information of the test set is leaked since most sequence decomposition methods are not strictly causal. Future information leakage usually causes considerable overestimation of model prediction accuracy. A correct method for utilizing sequence decomposition techniques without data leakage is using the sequences before the time of each test data point to compute decomposed subsequences for both training and test. In this way, the model is completely agnostic to future information.

While the direction being pursued is plausible, our systematic investigation suggests it may be misguided. Our findings uncover a frustrating issue that the effectiveness of signal decomposition methods is severely overestimated, probably due to the inconspicuous information leakage introduced by improper data processing practices that conduct signal decomposition on both past and future data (Figure 1). Given the powerful function-fitting ability of deep learning methods, even subtle and implicit leakage of future information can lead to our hallucination of prediction accuracy enhancement. Unfortunately, the inability to use leaked information in practical contexts may yield severe model performance degradation and introduce unrecognized risks to scientific research and engineering endeavors. If not corrected promptly, such a course could lead to potential misleads and resource waste in both academia and industry, as the painful lesson learned from "cardiac stem cells" [34]. Our revisiting is expected to help calibrate researchers' estimation of the performance of signal decomposition-enhanced time series forecasting AIs, meanwhile providing valuable guidance and insight for engineering trouble-shooting, technological evolution, and scientific progress.

Our work is also intended to inspire the research community to scrutinize certain well-established technological paradigms and proactively identify those neglected pitfalls.

## 2. Result

We conduct extensive experiments on six datasets in different domains, including significant wave height, wind speed, humidity, solar power, air pressure, and temperature (see details in Supplementary Methods). We select three representative signal decomposition methods, i.e., Empirical Mode Decomposition (EMD) [35], Discrete Wavelet Transform (DWT) [36], and Singular Spectrum Analysis (SSA) [37]. We use two different data processing procedures: (1) decomposition on all time steps, which introduces concealed future information leakage, and (2) decomposition on strictly restricted causal decomposition, which can only utilize past observations.

We test the results of the naive persistence baseline, the vanilla LSTM model, and LSTMs combined with leaked or non-leaked decomposed sequences (Figure 2). On all six datasets, the vanilla LSTM consistently achieves lower errors than the persistence strategy, showing its expected predictability. The prediction errors decrease at an abnormal scale when incorporating decomposed signals with inherent future label leakage, while the errors are calibrated when using the non-leaked data. For all methods on all datasets, a slight performance degradation is observed compared with the vanilla LSTM when we incorporate decomposed sequences without a glance at future data points. SSA shows the most serious leakage problem among different decomposition methods, i.e., the prediction errors reduce up to more than 90% on all datasets.

These phenomena are unexpected but reasonable, since the sequences decomposed on the entire series encode rich clues about future trends, such as forthcoming peaks and valleys. Thus, the models learned and tested on sequences with potential data leakage usually show superior but inauthentic results. Moreover, the effectiveness of signal decomposition in natural time series prediction is challenged by the results, since it fails to boost the performance of the basic LSTM model. The above findings show some possible contradictions to some prior studies [38,

39, 40]. There is a small but non-negligible possibility that, this data leakage issue exists in a broad range of research in the natural sequence area, causing an over-estimation of the effectiveness of signal decomposition in predictive problems.

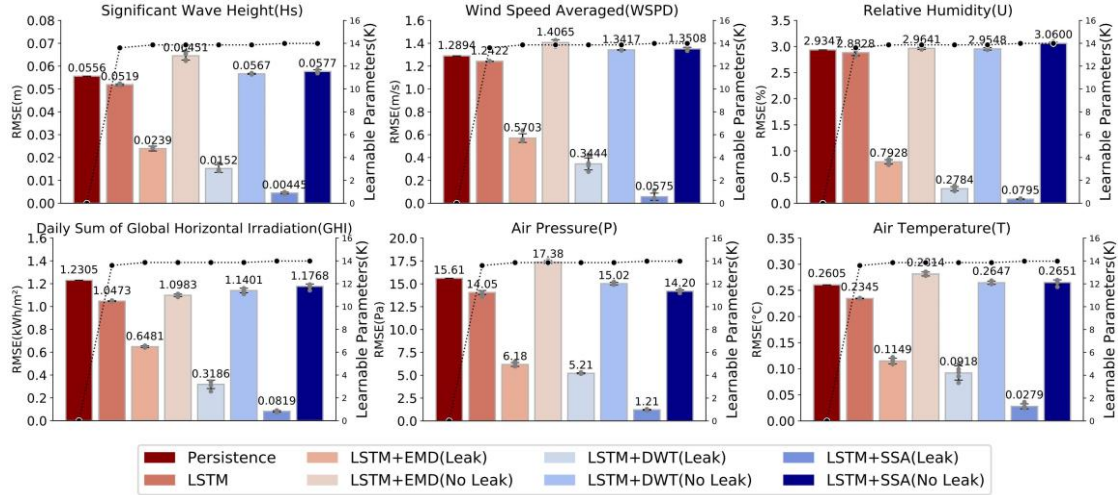

Figure 2: Influence of potential data leakage in signal decomposition on prediction performance. Error bars are standard deviations of five repeated experiments. Comprehensive analysis is conducted on six datasets with different physical time series and different signal decomposition methods, including Empirical Mode Decomposition (EMD) [35], Discrete Wavelet Transform (DWT) [36], and Singular Spectrum Analysis (SSA) [37]. When introducing decomposed time series with future information encoded, the prediction errors consistently decrease at an unprecedented scale (e.g., 91.4% for SSA on the significant wave height dataset). The actual debiased performance with a strictly causal decomposition process substantially degrades, indicating rather limited and even negative impacts of decomposed signals on model predictability (the performance differences between leaked and non-leaked versions are significant in two-sided t-test, $p < 0.001$). Since the number of learnable parameters remains similar, it is highly suspected that data leakage may lead to a hallucination of the effectiveness of decomposed sequences.

We also reproduce this worrying issue in different time series prediction models. By comparing the vanilla MLP, LSTM, and Transformer models and their variants combined with decomposed sequences (Figure 3), we verify the universal impact of data leakage on all compared models. Although these models have different architectures, model parameter capacities, and computation processes, the implicit leakage of future data can be consistently captured by them. Even the time-agnostic MLP model can achieve rather low errors when the leak occurs. This result reveals that, a large family of machine learning models can be influenced by the improper data processing method.

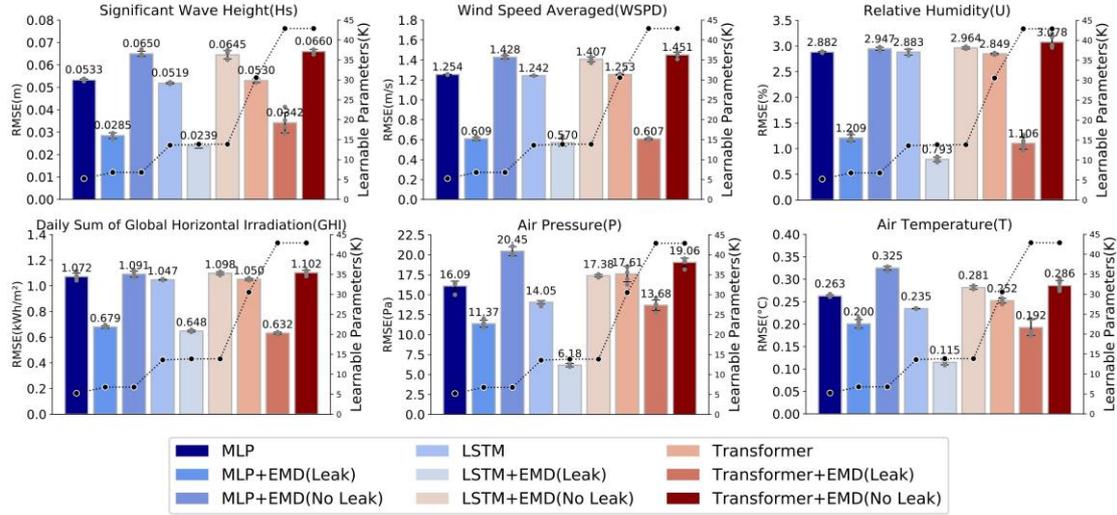

Figure 3: The impact of data leakage on various time series prediction models. Error bars are standard deviations of five repeated experiments. We test three canonical neural network architectures, i.e., multi-layer perceptrons (MLP), long short-term memory (LSTM) [41] network, and Transformer [42], assisted by the represented signal decomposition method EMD. Although these models are diverse in their characteristics and model parameter sizes, consistent mismatches between models learned on leaked and non-leaked data are observed ($p < 0.001$ in two-sided t-test). It indicates a universal problem in the data processing procedure of signal decomposition-based time series prediction research.

To further analyze the impact on the prediction behaviors of time series prediction models, we visualize the real observations and the prediction curves on different datasets (Figure 4). We use LSTM as the basic model and EMD as the decomposition method. The prediction curves with data leakage closely match the trends of real observations, which is consistent with the quantitative analysis above. However, for the model tested on non-leaked data, the predictions have lower sensitivities to frequent or drastic fluctuations, as shown in subplots A, B and D. This probably indicates a reason for the ineffectiveness of sequence decomposition in these cases, i.e., the frequency domain signals extracted from the past observations may enforce the model to follow past trends rather than discover upcoming patterns. The overfitting problem of these frequency signals makes models less sensitive to the quick changes in the time domain, since these fluctuations correspond to outliers in the frequency domain that are missing in past training data. Thus, the real effectiveness of signal decomposition in time series prediction is not as salient as we expect.

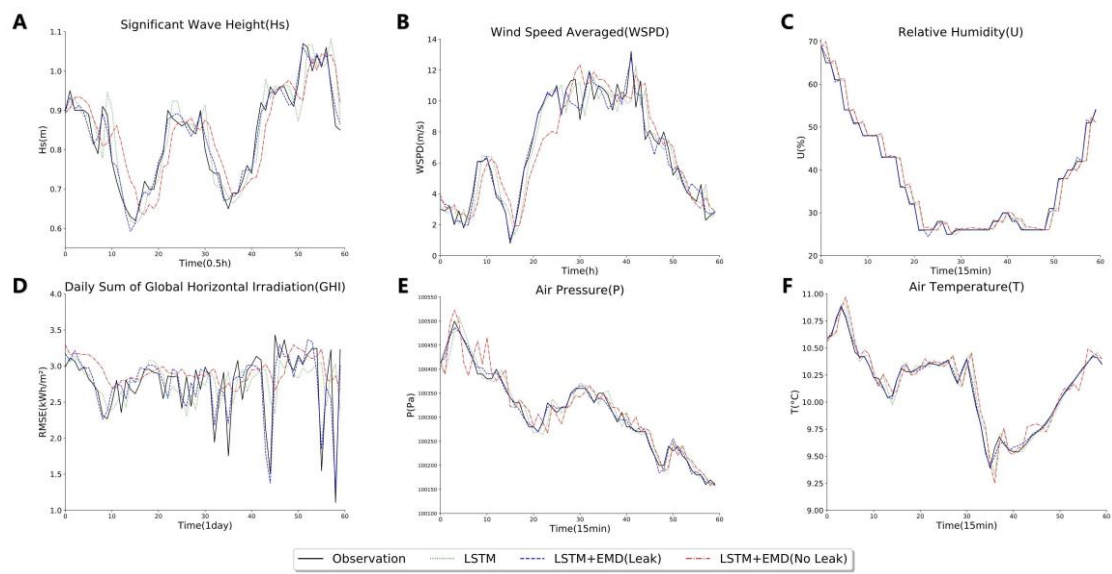

Figure 4: The groundtruth curves and predicted curves of a randomly selected sample on different datasets. We compare the LSTM model with raw sequences only and EMD sequences with or without leakage. We observe different prediction patterns between leaked and non-leaked versions. For samples with frequent or drastic fluctuations (e.g., subplot A, B, and D), the non-leaked EMD version shows lower sensitivity and higher biases compared with the original LSTM model. It reveals a phenomenon that introducing extra frequency domain signals may weaken the model's capability of modeling strong fluctuations in the time domain, possibly due to the overfitting of past frequency domain data.

Finally, we evaluate the impact of different components in the family of decomposed sequences. We take EMD as an exemplary method for analysis (see Supplementary Figures E.17 and E.18 for results of other decomposition methods). We add one of the different components into the LSTM model and compare its performance change (Figure 5). We uncover that the abnormal error decrease is majorly brought by the high-frequency components (see the spectrum in Supplementary Figures E.9, E.10, E.11, E.12, E.13, E.14), and the first sequence has the most salient contribution. It implies that the model mainly overfits the fluctuation patterns encoded by high-frequency components, which have direct correlations with the actual values of future data observations. Based on this finding, we can explain the phenomenon that SSA leads to greater leakage impacts than EMD (Figure 2), since SSA draws the global picture of signals in the frequency domain while EMD only focuses on the patterns of local minima and maxima. This may further indicate that the improper data processing method may cast larger impacts on signal decomposition methods with heavier frequency information dependencies.

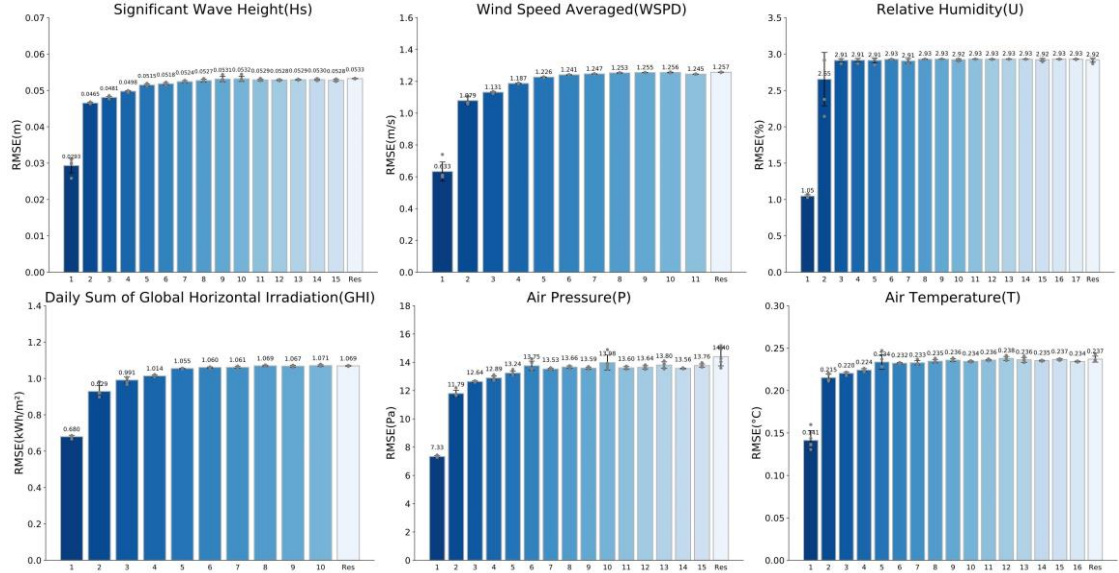

Figure 5: Impact of residual and different Intrinsic Mode Functions (IMFs) on information leakage. Error bars are standard deviations of five repeated experiments. We add one of the IMFs in EMD or the residual part into the LSTM model in each experiment. Sequences with smaller indexes contain higher frequency components. The results show that the first component with the highest frequency has the most salient impact on the abnormal performance change. A few high-frequency components show major effects and the rest low-frequency ones do not significantly contribute to the label leakage. It shows the subtle local trends of time series play more important roles in time series prediction and the leakage of such clues causes more misleading conclusions.

## 3. Discussion

Time series prediction is a pivotal problem across various scientific research domains. With the accumulated studies and verifications from different scenarios, it seems that a consensus has been achieved that signal decomposition techniques are critical complementary to prediction models in terms of embedding physical-world knowledge. Unfortunately, after thorough analysis and scrutinization, we uncover an unexpected matter that the effectiveness of signal decomposition is exaggerated due to an inconspicuous type of data leakage. We confirm that, decomposing the time series on the entire data including test splits introduces considerable future information leakage, even if we incorporate the decomposed data points before the test time only. Such leakage causes a universal and dramatic performance overestimation of various models in various scenarios. We also uncover a phenomenon that high-frequency patterns encoded with future information majorly account for the label leakage, since they are probably more informative for forecasting short-term future trends. Our work whistleblows an alarming issue in the time series prediction area that a well-established practice seems to be questionable and needs comprehensive re-examination.

The act of revisiting scientific findings holds immense value and significance as time series prediction is a fundamental aspect of many fields. Unfortunately, unintended mistakes have led to scientific errors, missed opportunities, and economic and social losses. This highlights the need to improve the systematic and reproducible nature of scientific research and encourage healthy and sustainable development of the scientific research ecology and the linkage between science and society. We aim to use our work as a flag in the process of accumulating scientific knowledge to champion these ideals and address problems hidden in other fields.

Our study has several limitations. Firstly, we cannot ensure that we have exhaustively reviewed all relevant literature and available approaches to investigate the existing issues. Secondly, there are diverse decomposition methods that can result in varying effects when applied in practice. We have only compared them based on typical and mainstream methods, and thus, other scenarios may exist. Our retrospective study aims to remedy past errors, and we cannot directly identify and correct existing flaws in actual systems. Therefore, it is necessary for researchers and technical experts to collaborate to minimize the impact of existing and potential issues.

## 4. Methods

Here we introduce the detailed experimental protocol of our analysis (Figure 1). Given a time series prediction dataset $D$ with $M$ independent samples, we divide each sample into two parts chronologically for training and test. For each sample $x \in D$, we denote the subsequence in the training and test sets as $x_t = [x_1, x_2, ..., x_P]$ and $x_v = [x_{P+1}, x_{P+2}, ..., x_Q]$, respectively, where $P$ and $Q - P$ are their sequence lengths. We then use signal decomposition methods such as EMD, DWT, and SSA to obtain the decomposed sequences of the training sequence $x_t$, which are denoted as $X_d = [x_t^1, x_t^2, ..., x_t^K]$, where $K$ is the number of decomposed components. In a typical method [33] for incorporating decomposed signals, the raw sequence $x_t$ and the series of decomposed sequences $X_d$ are concatenated together to form the processed data $X_p$.

The time series prediction model receives the above sequences as the input. It can be implemented by various architectures, such as traditional machine learning methods (e.g., SVM [43]) and deep learning models like MLP [44], CNN [45], LSTM [46], and Transformer [47]. Denote the prediction window of this model as $W$, then the model conducts prediction on $X_p$ in a sequential way as follows:

$$\begin{aligned} \hat{x}_{W+1} &= f(X_p[1:W]; \theta), \\ &\cdots \\ \hat{x}_{P-1} &= f(X_p[P-1-W:P-2]; \theta), \\ \hat{x}_P &= f(X_p[P-W:P-1]; \theta), \end{aligned} \quad (1)$$

where $f(\cdot; \theta)$ represents the non-linear mapping function learned by the model with parameters $\theta$, $[i:j]$ stands for the slicing operation from the $i$-th to the $j$-th steps, and $\hat{x}_i$ means the predicted value at the $i$-th step. By comparing the predicted values with the real observations, we then compute the loss function $\mathcal{L}$. Taking the commonly used mean squared error [48] as an example, the loss function on the sample $x$ is computed as follows:

$$\mathcal{L} = \sum_{i=1}^{P-W} (x_{W+i} - \hat{x}_{W+i})^2. \quad (2)$$

The model parameters $\theta$ are therefore updated by minimizing $\mathcal{L}$ via gradient descent algorithms (e.g., Adam [49]). For more efficient and robust training, we also use the batch model training strategy by simultaneously optimizing the model on multiple samples, until the model converges. For easy model hyperparameter tuning, we randomly select 10% of training samples as the validation set.

In the test phase, we use the model to predict the observations in the test split. We consider two types of data preparation methods. The first one is the leaked version, where the entire sequence $x$ is used for signal decomposition. we denote its decomposed sequences (with $Q$ observation values) as $\hat{X}_d$. For each prediction step $i$, we use the sliced subsequences of $\hat{X}_d$ as the model input for prediction as follows:

$$\hat{x}_{P+i} = f(\hat{X}_p[P+i-W:P+i-1];\theta), \qquad (3)$$

where $\hat{X}_p$ means the combination of the raw sequence $x$ and $\hat{X}_d$. Since most signal decomposition methods like EMD and SSA are not strictly causal in theory, the decomposed sequences in fact contain the information on future observations, even though we only consider the data points before the prediction cutoff step.

By contrast, the second one is a non-leaked version, where only the data points before the prediction time are used for signal decomposition. For the $i$-th prediction step, we use the truncated sequence $x[1:P+i-1]$ to compute the decomposed sequences without future information leakage (denoted as $\hat{X}_d$). It is further combined with the original sequence $x$ as the model input to infer the estimated value as follows:

$$\hat{x}_{P+i} = f(\tilde{X}_p[P+i-W:P+i-1];\theta), \qquad (4)$$

where $\tilde{X}_p$ means the combination of the raw sequence $x$ and $\hat{X}_d$ without future information leakage. In this way, we ensure that the model can only be aware of the information before the time for prediction, thereby no label information is leaked. If there is a sufficiently significant difference between the prediction accuracy obtained by the two data processing methods, we can confirm that the improper data processing procedure indeed brings unwanted overestimation of the effectiveness of signal decomposition-based time series prediction methods.

## Data Availability

The datasets involved in this study are all publicly available. The significant wave height (Hs) dataset is available at http://cdip.ucsd.edu/offline/wavecdf/wnc browse.php?ARCHIVE/150p1/150p1 historic. The wind speed averaged (WSPD) dataset is available at https://www.ndbc.noaa.gov/. The relative humidity (U) dataset is available at

https://www.kaggle.com/datasets/l3llff/electrical-grid-power-mw-20152021. The daily sum of global horizontal irradiation (GHI) dataset is available at https://solargis.com/products/evaluate/useful-resources. The air pressure (P) and temperature (T) datasets are available at http://maps.nrel.gov/wind prospector. All experiments and implementation details are described in sufficient detail in the Methods section or in the Supplementary Information.

# Appendix A. Supplementary Methods

## Appendix A.1. Dataset

We use 6 datasets for natural time series prediction in different scenarios. Their details are listed below.

- Significant wave height (Hs) dataset, which comes from the Coastal Data Information Program (CDIP) (http://cdip.ucsd.edu/offline/wavecdf/wnc browse.php?ARCHIVE/150p1/150p1 historic). It was collected on-site at the designated location (34.142 N, 77.710 W) of UCSD (University of California, San Diego), and the time range was from January 1, 2017, to December 31, 2020, i.e., a total of four years. Significant wave height data has a time resolution of 30 minutes and a total of 70,128 points.

- Wind Speed Averaged (WSPD) dataset, which comes from the measured data of National Data Buoy Center's East Coast site (Station 43.525 N, 70.141 W, https://www.ndbc.noaa.gov/). WSPD includes the average wind speed data of a total of 32,158 points from January 1, 2002 to December 31, 2005 with a time resolution of one hour.

- Relative humidity (U) dataset, which comes from the Kaggle website (https: //www.kaggle.com/datasets/l3llff/electrical-grid-power-mw-20152021?resource=download), is the measured data at a power plant in Germany. The denotation "U" represents the relative humidity at a height of 2 meters above the surface. The dataset covers the time period from December 31, 2014 to July 8, 2021, with a time resolution of 15 minutes, resulting in a total of 228,526 data points.

- The daily sum of global horizontal irradiation (GHI) data, which comes from Solargis website (https://solargis.com/products/evaluate/useful-resources). GHI is the site data generated at the location Plataforma Solar de Almeria, Spain (coordinates: 37.094 N, 2.360 W, elevation: 497.0m a.s.l.). The GHI contains data for a total of 9,952 points from January 1, 1994, to March 31, 2021, with a temporal resolution of one day.

- Air pressure (P) dataset is a simulated one from National Renewable Energy Laboratory (NREL) (http://maps.nrel.gov/wind prospector), which measures the air pressure at an elevation of 100 meters at the west coast of the United States (41.812 N, 124.317 W). It includes four years of air pressure from January 1, 2014 to December 31, 2017, with a time resolution of 15 minutes and 140,160 data points.

- Air temperature (T) dataset is also a simulated one from the same source as air pressure (P)(http://maps.nrel.gov/wind prospector). It measures the air temperature at the same elevation and location, with the same time resolution of 15 minutes and the number of data points of 140,160.

We present exemplary sequence diagrams (Supplementary Figure E.6) and detailed statistics (Supplementary Table D.1) of the above 6 datasets.

## Appendix A.2. Experimental Setup

All data used in this study are processed using the information from the previous 12 steps to predict the subsequent step. The datasets are divided by time into a training set and a test set with a ratio of 0.75 to 0.25. Prior to training, the data are normalized using the MinMaxScaler [50] technique. For both the MLP and LSTM models, the maximum training epoch is 1,000 and the batch size is 32. The optimizer we use is Adagrad [51] for MLP and Adam [49] for LSTM, and the learning rate is 0.0001. We set the epoch patience for early stopping as 30, i.e., the training will be terminated if the validation loss does not improve within the last 30 epochs. The MLP model contains three non-linear layers. The LSTM model consists of two LSTM layers followed by two fully-connected layers. The Transformer model we use has four Transformer encoder layers to extract features from the input sequence. The input feature dimension is set to 32. By using sinusoidal positional encoding, the model encodes the input sequence and adds it to the input so that positional information can be captured. In each Transformer encoder layer, the multi-head self-attention module has 8 attention heads and their output dimension is 32. The outputs from the Transformer encoders are aggregated using average pooling to obtain the hidden representation of the entire sequence. A fully connected layer with 128 neurons and ReLU [52] activation function maps it to the predicted value. The optimizer employed is Adam, and a custom learning rate scheduler [42] is utilized to dynamically adjust the learning rate for better convergence. For the SSA method, we set the number of decomposed sequences to 3. For all models, the loss function and validation metrics are evaluated using the Mean Absolute Error (MAE). In the test phase, we use Mean Squared Error (MAE) as the main metric and include Mean Absolute Percentage Error (MAPE) as well as the coefficient of determination R2 as supplementary metrics.

## Appendix A.3. Methods for Sequence Decomposition

In this section, we introduce the three popular sequence decomposition methods used in our experiments.

Empirical mode decomposition (EMD) [35], a method based on the Hilbert Huang transform (HHT) that has been successfully applied to various time-series analysis tasks. It has the following advantages: (1) suitable for nonlinear and non-stationary signals; (2) unlike wavelet transform (WT) or Fourier transform (FT), HHT has complete adaptivity by introducing intrinsic mode functions (IMFs), which do not require the pre-selection of basis functions. To successfully decompose into IMFs, two constraints are required. The first is that the number of extrema and zero crossings is either equal or differs by no more than one. The second is that the mean value of the upper envelope formed by local maxima and the lower envelope formed by local minima is zero at any given time. During the iterative process of EMD, after obtaining the first IMF, it is subtracted from the original signal to derive a new residual signal. Subsequently, the same iteration is repeated for the residual signal until all the IMFs and a final residual (Res) component are obtained. In our experiments, we obtain different numbers of IMFs and a residual subsequence (Supplementary Figures E.9, E.10, E.11, E.12, E.13, and E.14), as well as their relative frequency spectrum obtained by the Fast Fourier Transform method.

Discrete Wavelet Transform (DWT), a process of transforming a signal into simpler signals using wavelets. It converts the signal into sets of frequency components, each of which is characterized by a particular scale or frequency. It is used for various applications such as image compression, denoising and data analysis. DWT decomposes the signal into different levels of resolution, providing a multi-scale representation of the original signal by representing it at different frequencies and scales instead of just one size fits all approach. In our study, we chose the db5 wavelet basis function in Daubechies, and the original time series are transformed by discrete wavelet transform [36] to obtain two subsequences, i.e., an Approximate component and a Detailed component (Supplementary Figure E.15).

Singular spectrum analysis (SSA), which is a time series decomposition technique that provides an effective way of decomposing a time series into its constituent elements based on singular value decomposition (SVD). In SSA, each component of the time series is represented by its own "eigenmode" or "pattern". In our study, we utilize singular spectrum analysis [37] to decompose the original time series of each dataset into three subsequences for experiments (Supplementary Figure E.16).

# Appendix B. Possible Clues of Information Leakage in Existing Literature

After extensive literature review and examination, we find a number of clues that indicate potential information leakage related to questionable data processing methods for sequence decomposition in time series prediction. The coverage of literature concerning this problem is far beyond our expectations, and here we name a few possible examples to show this worrying issue:

A recent work [38] uses the time-varying filter-based empirical mode decomposition technique and LSTM for monthly rainfall forecasting. From the original dataset processing diagram taken from this article (see Supplementary Figure E.7A), the decomposed sequences have 660 data points in total. As mentioned in the original paper content, the first 495 data points are used for training and the rest 165 data points are for test. In this way, the information of the test samples is leaked by the training samples.

Another work [39] combines wavelet transform and neural network for multistep lead-time prediction of significant wave height. From the original visualization results taken from this article (see Supplementary Figure E.7B), the decomposed sequences have the same number of data points as the original input sequences. The article mentions that the DWT operation is conducted on the full sequences. The first 75% of data points are used for training and the rest 25% samples are used for test. In a similar way, the information of the test set is encoded by the training samples.

In a study [40] that combines singular spectrum analysis and an improved fuzzy system for custom prediction, the same method is used to construct the training and test sets. It is mentioned that a time series of 1,000 data points is decomposed into 7 components. At the same time, we can clearly see from several charts (see Supplementary Figure E.7C) taken from the original article that compares the original and decomposed sequences, both of them have 1,000 data points. Since these sequences are directly divided into the training, validation and test sets, a similar leakage of information in the test set also exists.

From the above several studies on information leakage, it can be seen that the problem of information leakage has been around for a long time in a wide range of research communities [53, 54, 55, 56, 33]. Based on the verification of our experiments, there are huge gaps between their reported results and the expected results without any information leakage. In fact, we have documented many more studies with similar problems that lead to information leakage, resulting in an overestimation of the performance of sequence decomposition in time series prediction.

# Appendix C. Supplementary Experiments

## Appendix C.1. Impact of Different Components of DWT and SSA

In the main texts, we have shown the impact of different components of EMD. Here, we show the results for DWT and SSA (Supplementary Figure E.17 and E.18). For the DWT method, we add either the approximate component or the detailed component into the LSTM model and compare its performance change. For the SSA method, We add one of the different subsequences into the LSTM model. We find the test loss reduction of the detailed component of DWT is slightly larger than the approximate component. This is consistent with our observations when using EMD, where high-frequency components play more important roles in formation leakage. Nonetheless, both components have rich low-frequency and high-frequency signals, thereby their impacts on prediction performance changes are comparable. We also find that the first and the third subsequences of SSA show stronger leakage effects than the second one. Since all three sequences contain factors with different frequencies (e.g., high-frequency signals are not weak in all subsequences), it is probably because the first and third components contain more important high-frequency signals than the second one. Since SSA has a stronger data leakage effect than EMD and DWT according to our experiments, it would be possible that the high-frequency signals obtained by SSA leak future information more intensively.

## Appendix C.2. Experimental Results in All Metrics

Here we present the complete results of different experiments in all three metrics (Supplementary Table D.2). The phenomena reflected by different metrics are consistent.

## Appendix C.3. Experiments on Component Summation Methods

Different from taking all decomposed components as the input of a unified model, another popular method is to train separate models on different subsequences and sum up their predictions (Supplementary Figure E.8). Although this method is slightly different from the framework used in our main experiments, it has similar patterns according to the results of EMD-LSTM on the Hs dataset (Supplementary Table D.2 Supplementary Figure E.19). This shows that different usages of decomposed sequences may not alleviate the information leakage issue.

## Appendix C.4. Loss Curves in Model Training

We also show the training and test loss curves in different model learning epochs (Supplementary Figure E.20). In the training process of models with information leakage, the gaps between the training and test losses are very small, and even the test loss is lower than the training loss in some cases. On the contrary, the test loss of the non-leaked version is usually much larger than the leaked one, and the overfitting phenomenon is observed on all datasets when the number of epochs is too large. This further validates that abnormal performance improvements are brought by information leakage.

## Appendix D. Supplementary Tables

Table D.1: Statistics of the six datasets used in our experiments. Four datasets are real observations and two datasets are simulations.

|  | Hs(m) | WSPD(m/s) | U(%) | GHI(kWh/m$^2$) | P(Pa) | T(°C) |
|---|---|---|---|---|---|---|
| Sequence Length | 70,128 | 32,158 | 228,526 | 9,952 | 140,160 | 140,160 |
| Number of Training Samples | 52,596 | 24,118 | 171,394 | 7,464 | 105,120 | 105,120 |
| Number of Test Samples | 17,532 | 8,040 | 57,132 | 2,488 | 35,040 | 35,040 |

| | | | | | |
|---|---|---|---|---|---|
| Mean | 0.945 | 5.58 | 74.2 | 5.14 | 100.47k | 13.1 |
| Standard Deviation | 0.414 | 3.18 | 19.5 | 2.25 | 555 | 4.12 |
| Data Source Type | Observed | Observed | Observed | Observed | Simulated | Simulated |

Table D.2: The results of different models with different dataset processing methods. "Leak" indicates the data split method with future information leakage while "No Leak" indicates our causal data decomposition method. The results show that the leaked versions of all models show abnormal improvements over the original models without decomposed sequences. The performance decays when the information leakage is removed. The results in terms of different metrics exhibit consistent patterns.

| Model | Hs | | | WSPD | | | U | | | GHI | | | P | | | T | | |
|---|---|---|---|---|---|---|---|---|---|---|---|---|---|---|---|---|---|---|
| | MSE* | MAPE | $R^2$ | MSE | MAPE | $R^2$ | MSE | MAPE | $R^2$ | MSE | MAPE | $R^2$ | MSE | MAPE | $R^2$ | MSE | MAPE | $R^2$ |
| Persistence | 3.092 | 4.175 | 0.9812 | 1.663 | 30.28 | 0.8494 | 8.613 | 1.935 | 0.9772 | 1.514 | 22.79 | 0.6843 | 243.7 | 0.0106 | 0.9994 | 0.0679 | 1.352 | 0.9971 |
| MLP | 2.838 | 4.005 | 0.9828 | 1.573 | 30.15 | 0.8575 | 8.304 | 1.987 | 0.9781 | 1.149 | 23.58 | 0.7603 | 259.1 | 0.0112 | 0.9993 | 0.0690 | 1.397 | 0.9970 |
| MLP+EMD(Leak) | **0.813** | **2.210** | **0.9951** | 0.371 | 13.42 | 0.9664 | 1.466 | 0.824 | 0.9960 | 0.462 | 12.62 | 0.9037 | 129.5 | 0.0077 | 0.9997 | 0.0403 | 1.162 | 0.9982 |
| MLP+EMD(No Leak) | 4.224 | 4.873 | 0.9743 | 2.038 | 36.03 | 0.8153 | 8.688 | 2.227 | 0.9770 | 1.191 | 24.39 | 0.7517 | 423.2 | 0.0144 | 0.9989 | 0.1056 | 1.801 | 0.9953 |
| LSTM | 35.76 | 3.775 | 0.9836 | 1.543 | 30.15 | 0.8602 | 8.313 | 1.902 | 0.9780 | 1.097 | 22.92 | 0.7712 | 188.9 | 0.0096 | 0.9995 | 0.0550 | 1.176 | 0.9976 |
| LSTM+EMD(Leak) | 0.573 | 1.808 | 0.9964 | 0.326 | 11.86 | 0.9704 | 0.630 | 0.396 | 0.9983 | 0.420 | 11.50 | 0.9124 | 38.19 | 0.0042 | 0.9999 | 0.0132 | 0.598 | 0.9994 |
| LSTM+EMD(No Leak) | 4.165 | 4.781 | 0.9747 | 1.979 | 34.49 | 0.8207 | 8.786 | 2.367 | 0.9768 | 1.206 | 24.09 | 0.7484 | 302.1 | 0.0122 | 0.9992 | 0.0792 | 1.542 | 0.9966 |
| LSTM+DWT(Leak) | 0.233 | 0.902 | 0.9986 | 0.131 | 6.148 | 0.9896 | 0.079 | 0.174 | 0.9998 | 0.103 | 5.485 | 0.9786 | 27.18 | 0.0036 | 0.9999 | 0.0086 | 0.437 | 0.9996 |
| LSTM+DWT(No Leak) | 3.209 | 4.229 | 0.9805 | 1.800 | 30.36 | 0.8369 | 8.731 | 2.396 | 0.9769 | 1.300 | 23.33 | 0.7289 | 225.7 | 0.0105 | 0.9994 | 0.0701 | 1.434 | 0.9970 |
| LSTM+SSA(Leak) | 0.020 | 0.264 | 0.9999 | 0.004 | 1.158 | 0.9996 | 0.006 | 0.087 | 0.9999 | 0.007 | 1.425 | 0.9986 | 1.477 | 0.0008 | 0.9999 | 0.0008 | 0.128 | 0.9999 |
| LSTM+SSA(No Leak) | 3.325 | 4.390 | 0.9798 | 1.825 | 30.32 | 0.8347 | 9.365 | 2.723 | 0.9753 | 1.385 | 25.94 | 0.7111 | 201.5 | 0.0099 | 0.9995 | 0.0703 | 1.315 | 0.9970 |
| Transformer | 2.810 | 3.924 | 0.9829 | 1.570 | 31.03 | 0.8577 | 8.116 | 1.957 | 0.9786 | 1.103 | 22.91 | 0.7698 | 310.7 | 0.0113 | 0.9992 | 0.0636 | 1.323 | 0.9973 |
| Transformer+EMD(Leak) | 1.184 | 2.276 | 0.9928 | 0.368 | 13.10 | 0.9666 | 1.233 | 0.644 | 0.9971 | 0.399 | 11.21 | 0.9168 | 187.6 | 0.0085 | 0.9995 | 0.0372 | 1.014 | 0.9984 |
| Transformer+EMD(No Leak) | 4.355 | 4.810 | 0.9735 | 2.106 | 36.86 | 0.8092 | 9.483 | 2.598 | 0.9749 | 1.216 | 24.53 | 0.7465 | 363.4 | 0.0130 | 0.9990 | 0.0817 | 1.584 | 0.9965 |
| Component Summation | 0.483 | 1.477 | 0.9969 | 0.310 | 11.10 | 0.9716 | 0.783 | 0.710 | 0.9979 | 0.350 | 10.38 | 0.9249 | 178.9 | 0.0053 | 0.9995 | 0.0128 | 0.624 | 0.9993 |

*The MSE data of the Hs dataset has been enlarged by 1000 times (the unit is $m^2*10^{-3}$).

# Appendix E. Supplementary Figures

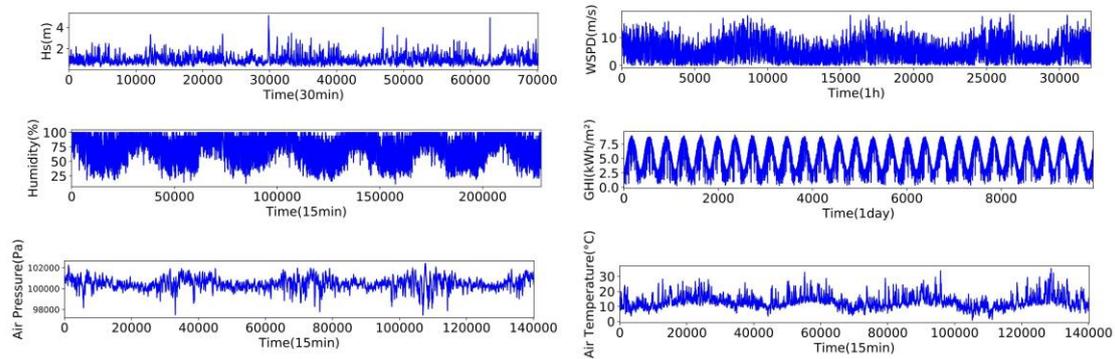

Figure E.6: Exemplary time series plots of the datasets used in our experiments. Different datasets are varied in their sequence lengths and resolutions.

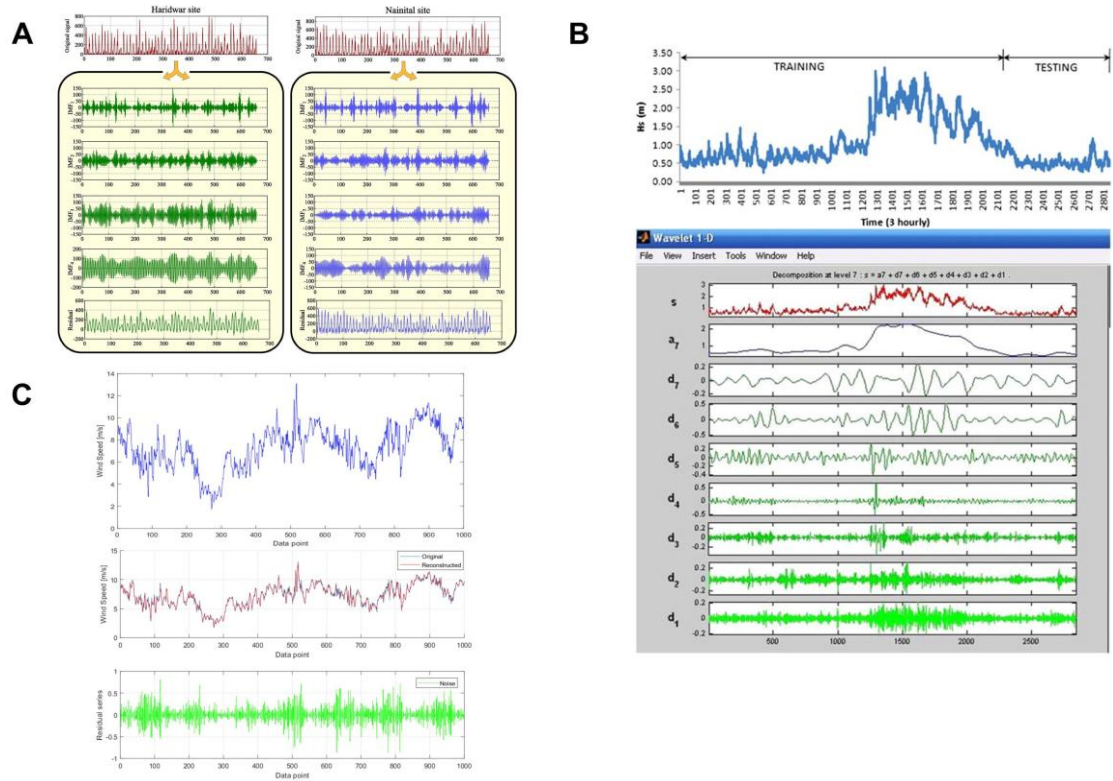

Figure E.7: The dataset processing charts taken from existing literature with possible information leakage [38, 39, 40]. (A) Each sequence with 660 data points is processed using EMD to obtain the IMFs and residual. (B) The sequences with about 2,800 points are processed by DWT to get the approximated component and multiple detailed components. The first 75% of points are used for training and the rest are for test. (C) The entire 1000-point sequence is decomposed into 7 components through SSA and then divided. All these figures show that the decomposed sequences with information leakage are used in model training.

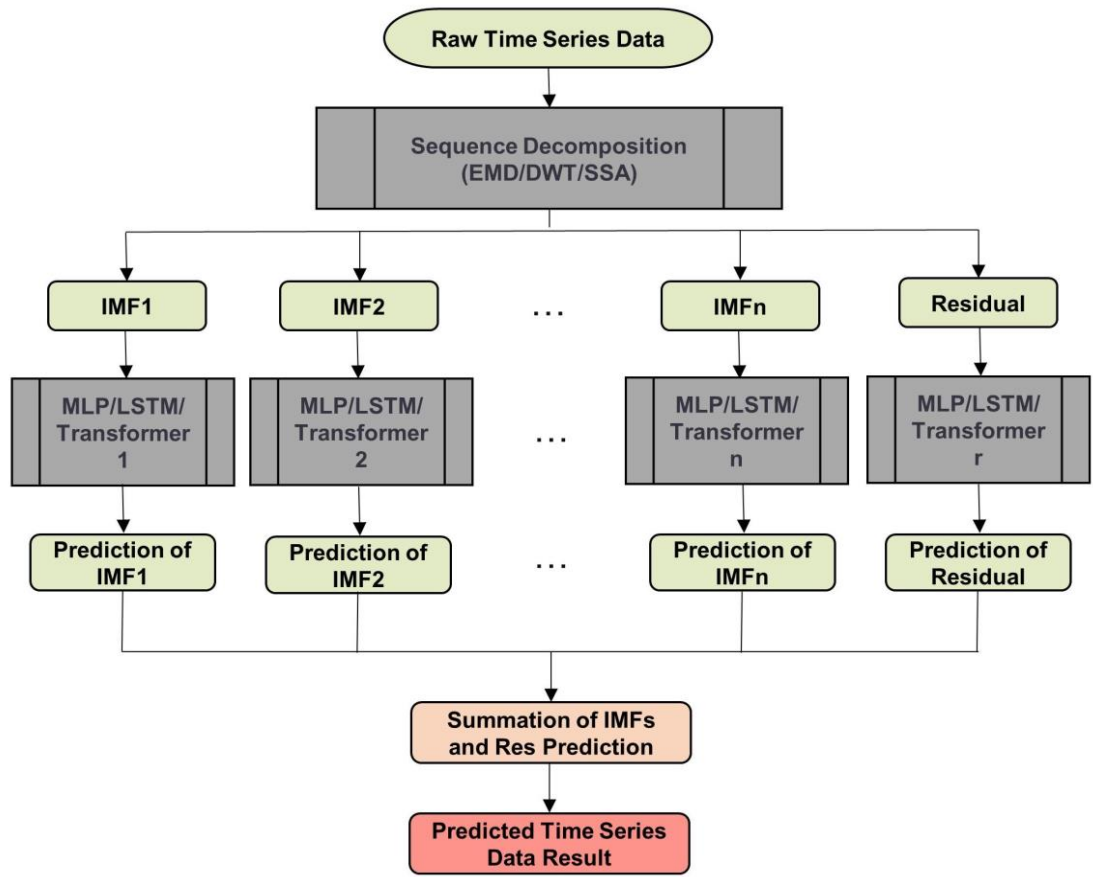

Figure E.8: The framework of decomposed competent summation. It uses separate models to predict based on different decomposed components, and sums up the predictions output by different models to obtain the final prediction. This framework also suffers from the information leakage problem since it takes the decomposed sequences as the model input.

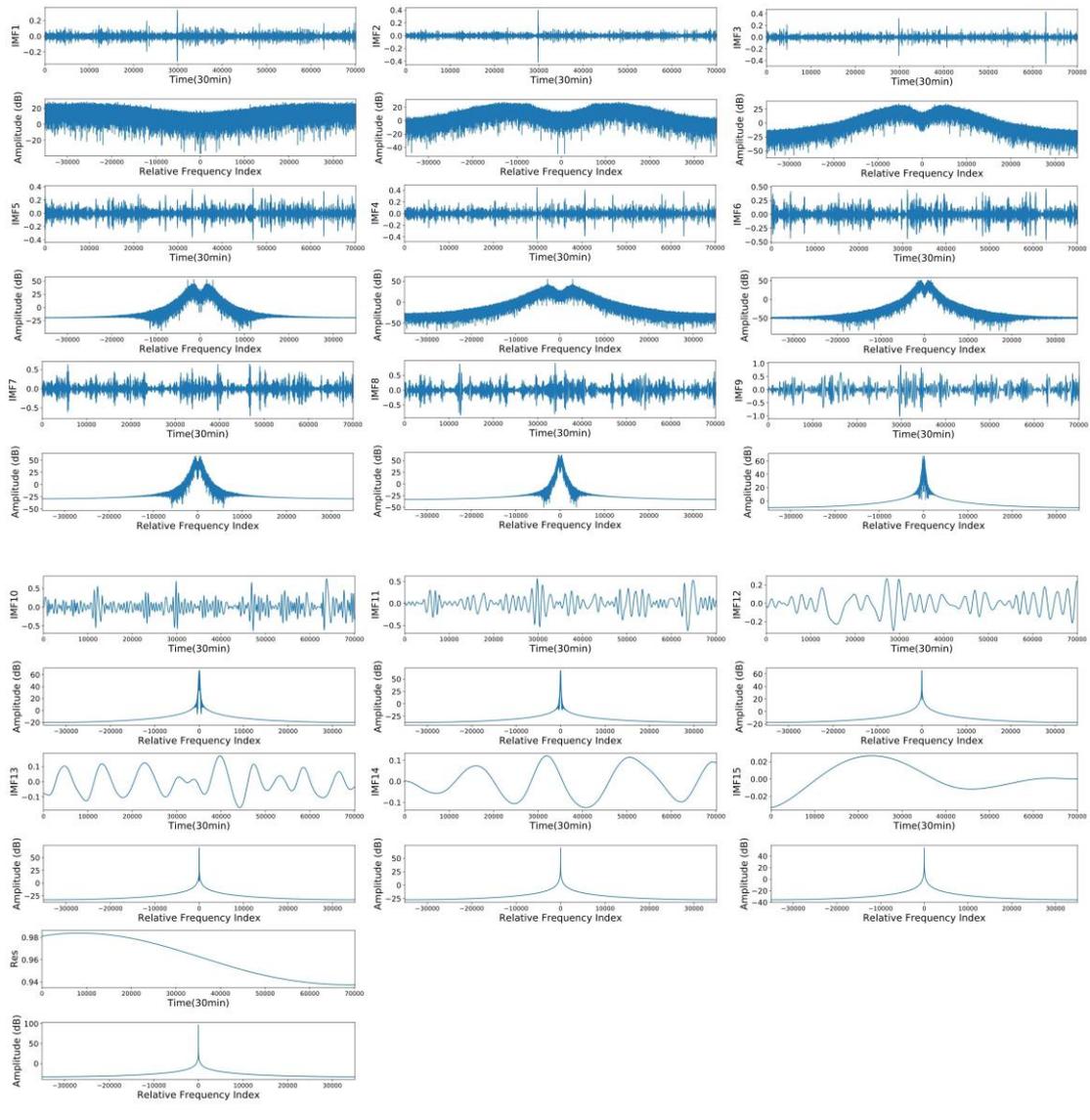

Figure E.9: The IMF and Res series obtained by EMD on the Hs dataset and their spectrums obtained by Fast Fourier Transform.

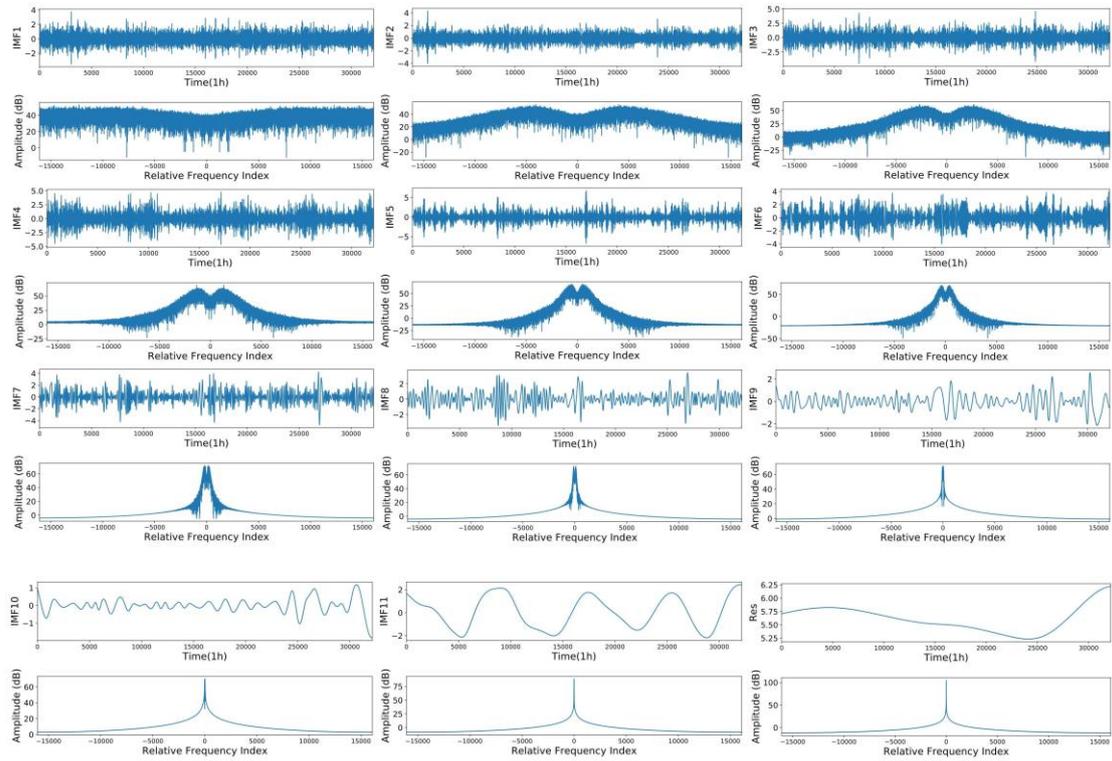

Figure E.10: The IMF and Res series obtained by EMD on the WSPD dataset and their spectrums obtained by Fast Fourier Transform.

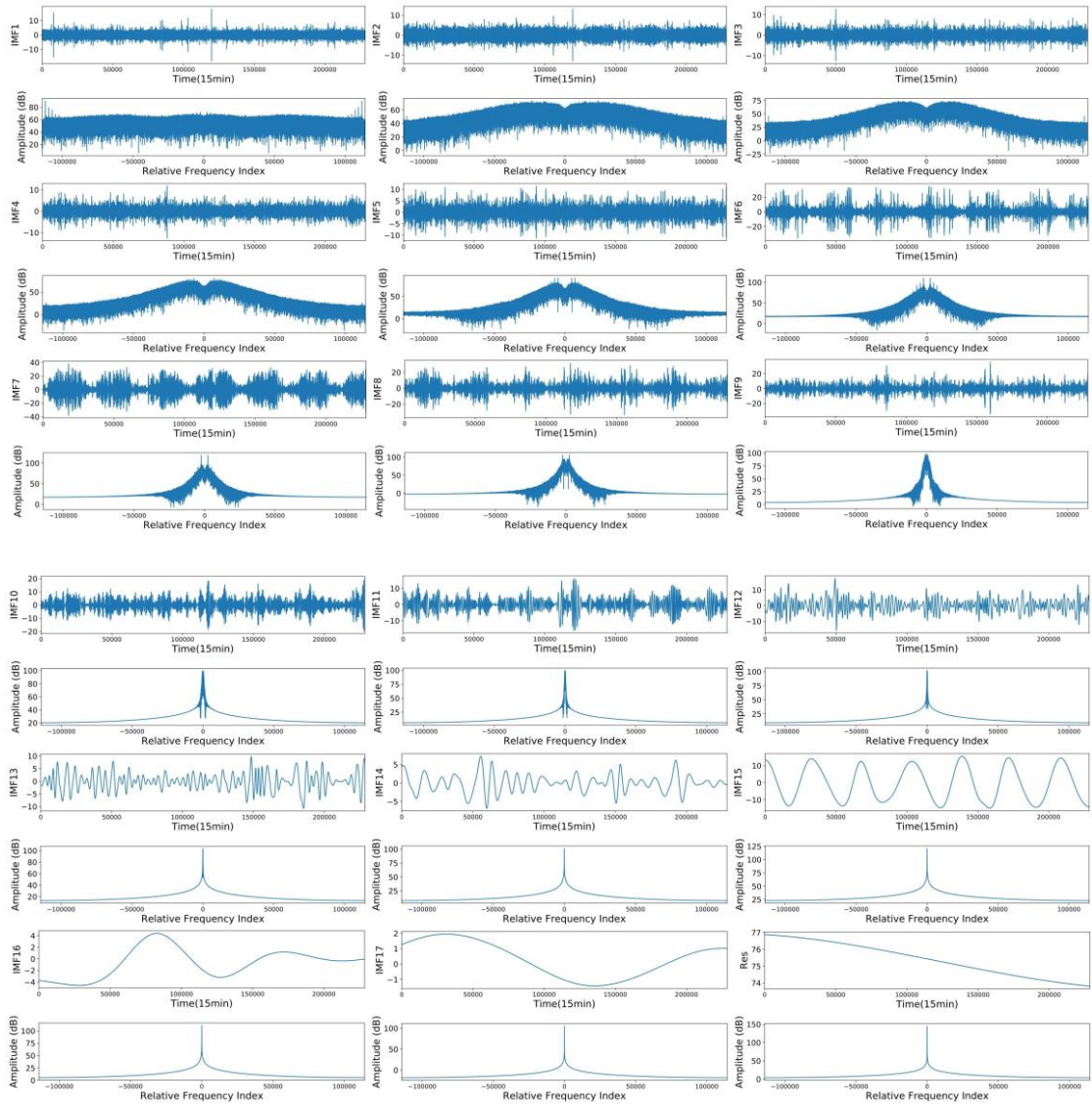

Figure E.11: The IMF and Res series obtained by EMD on the U dataset and their spectrums obtained by Fast Fourier Transform.

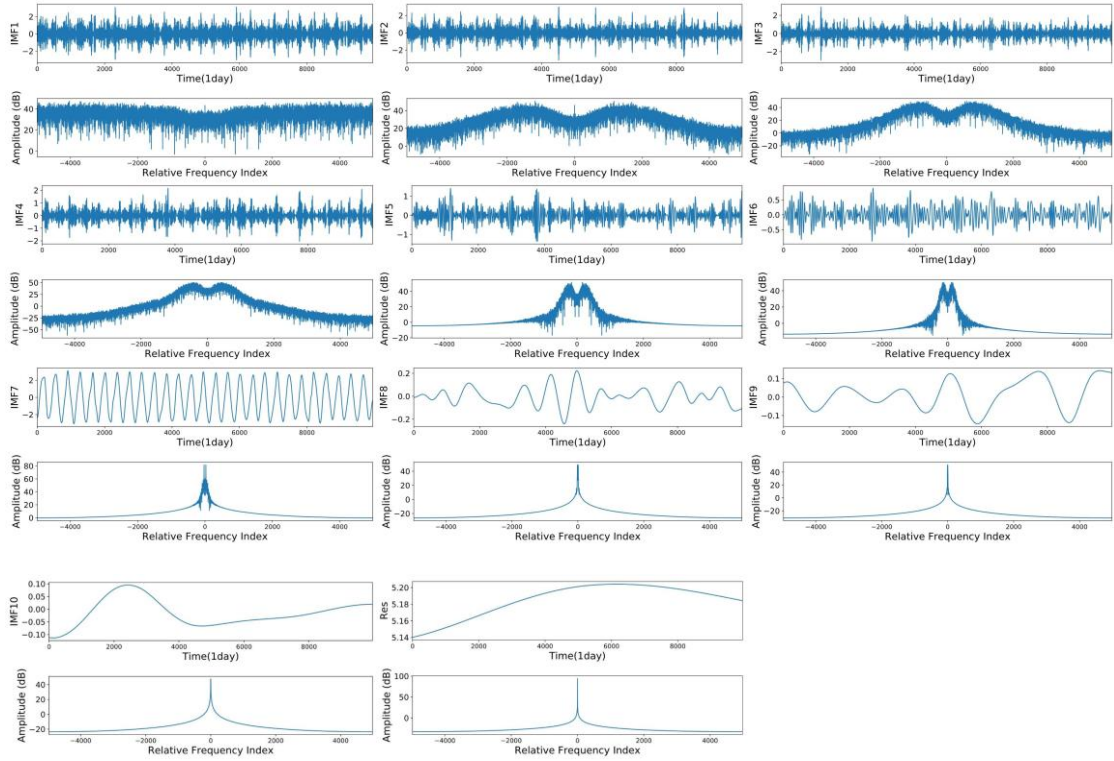

Figure E.12: The IMF and Res series obtained by EMD on the GHI dataset and their spectrums obtained by Fast Fourier Transform.

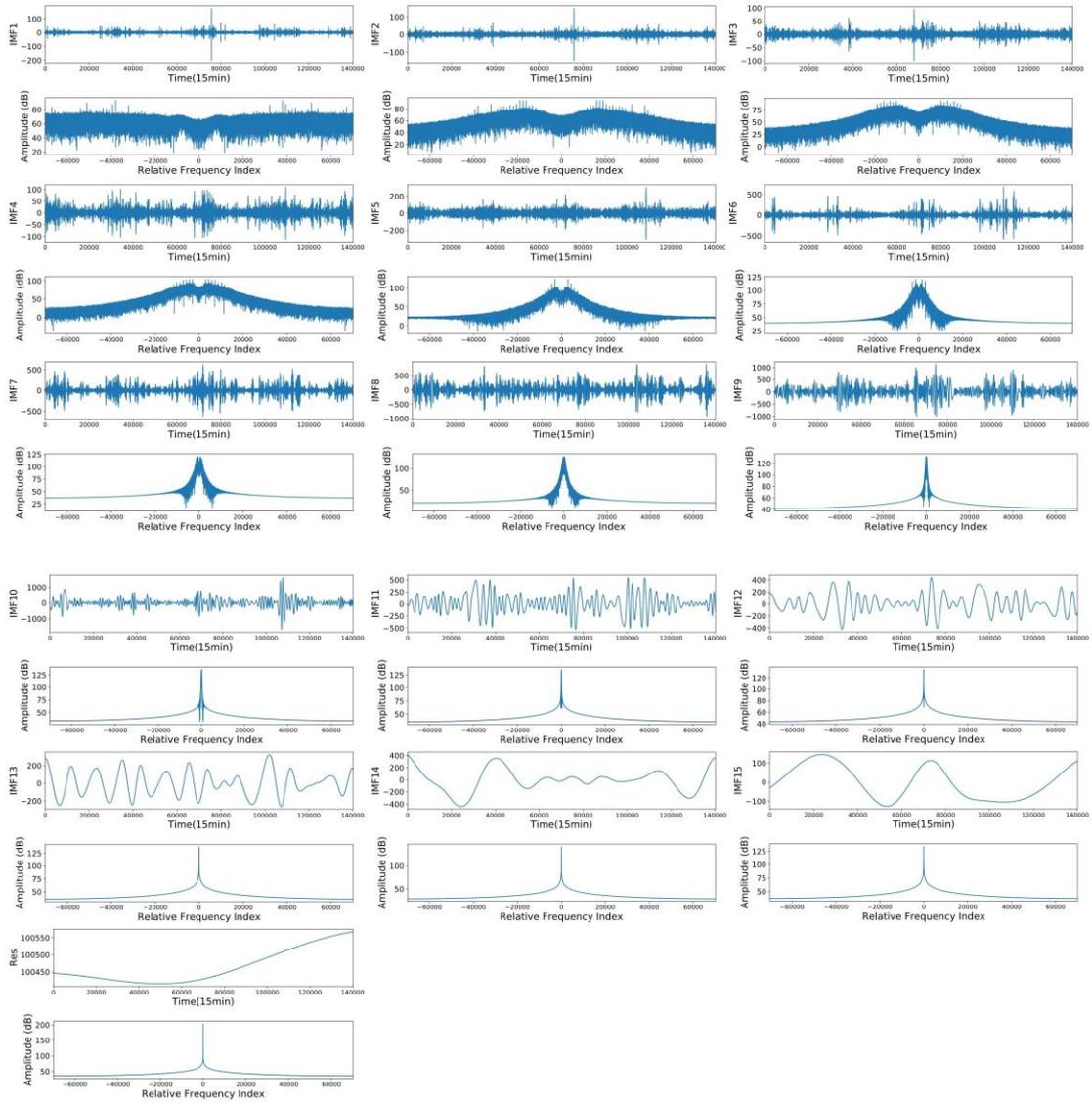

Figure E.13: The IMF and Res series obtained by EMD on the P dataset and their spectrums obtained by Fast Fourier Transform.

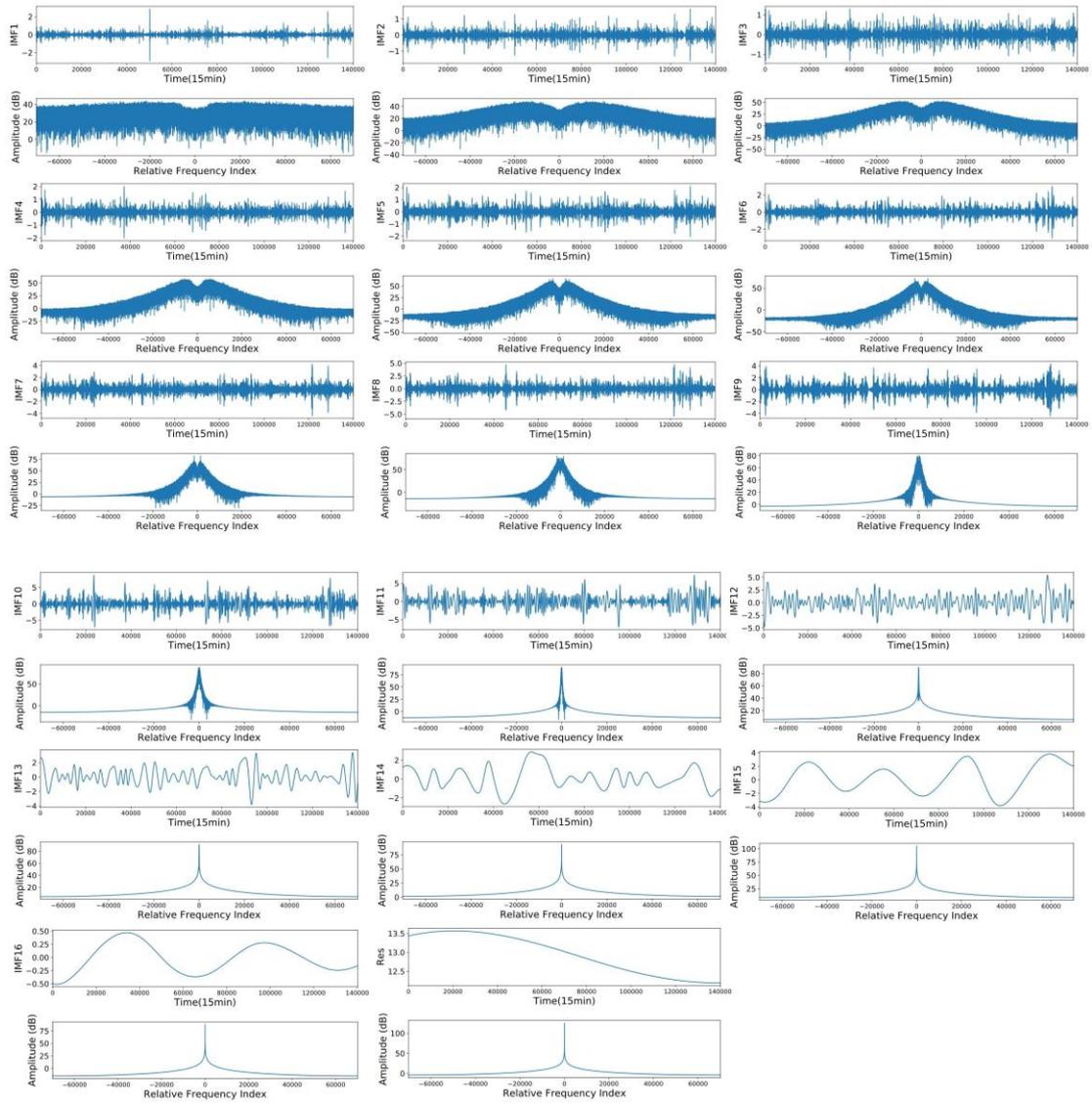

Figure E.14: The IMF and Res series obtained by EMD on the T dataset and their spectrums obtained by Fast Fourier Transform.

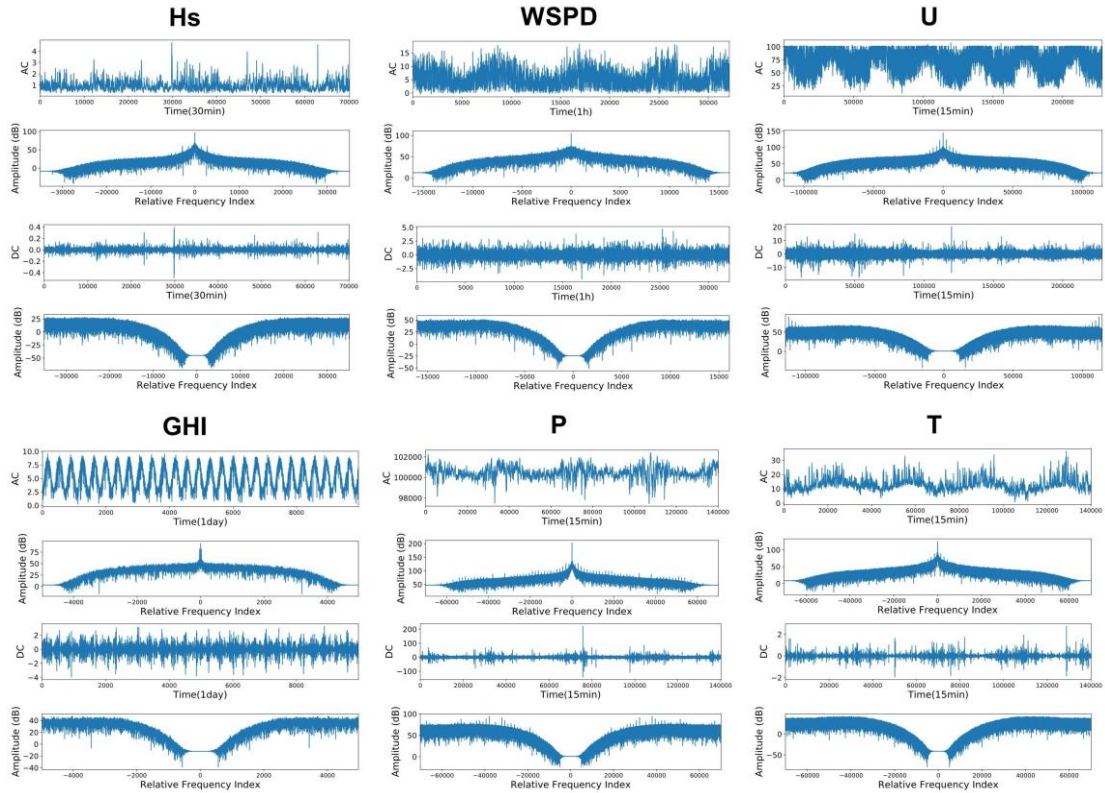

Figure E.15: The approximate component (AC) and detailed component (DC) sequences obtained by DWT on all datasets and their spectrums obtained by Fast Fourier Transform.

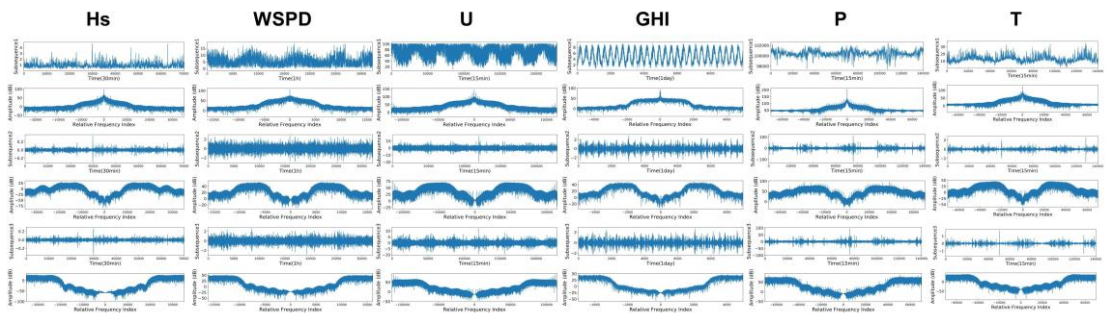

Figure E.16: The three subsequences obtained by SSA on the 6 data sets and their spectrums graphs obtained by Fast Fourier Transform.

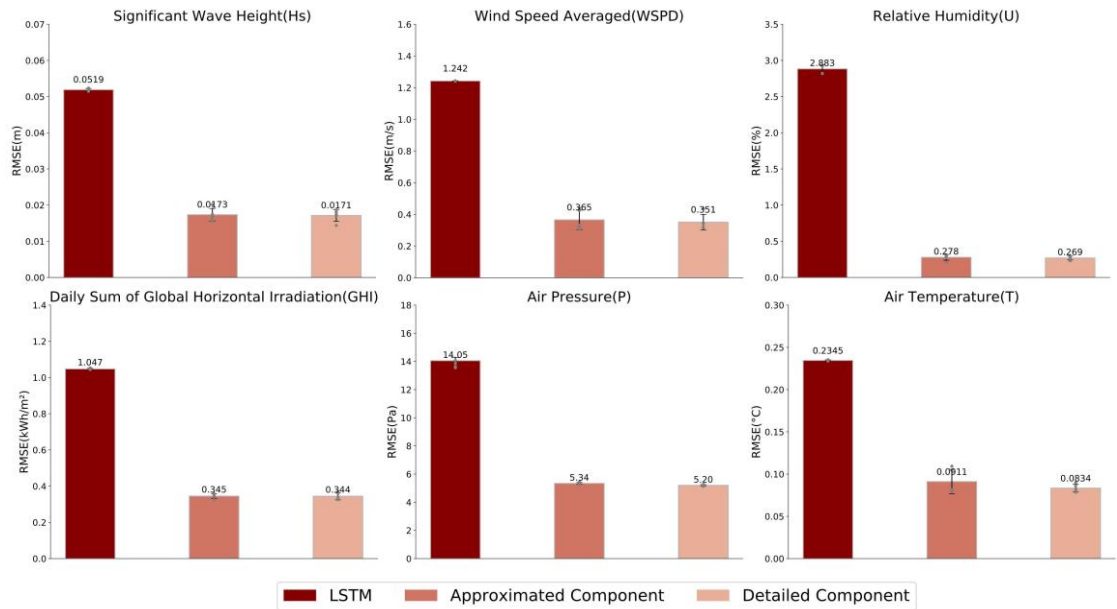

Figure E.17: Impact of the approximate component and the detailed component of DWT on information leakage. Error bars are standard deviations over five repeated experiments. We add one of two components to the LSTM model in each experiment. The approximate component has a lower relative frequency but is more similar to the original sequence, while the detailed component has a higher relative frequency. The influence of the detailed component is slightly larger than the approximate component on prediction error with information leakage.

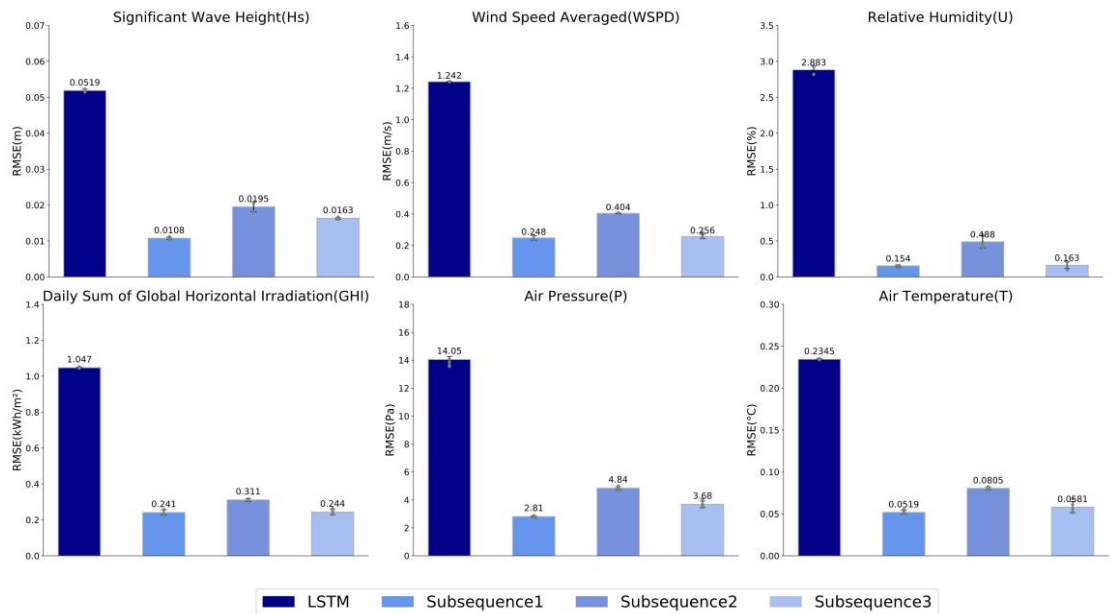

Figure E.18: Impact of different subsequences of SSA on information leakage. Error bars are standard deviations over five repeated experiments. We add one of the subsequences into the LSTM model in each experiment. The energy of sequences with larger indices is focused on higher frequencies. The influence of the second component is weaker than the first and third components. This may be because it has lower high-energy factors than the other two components.

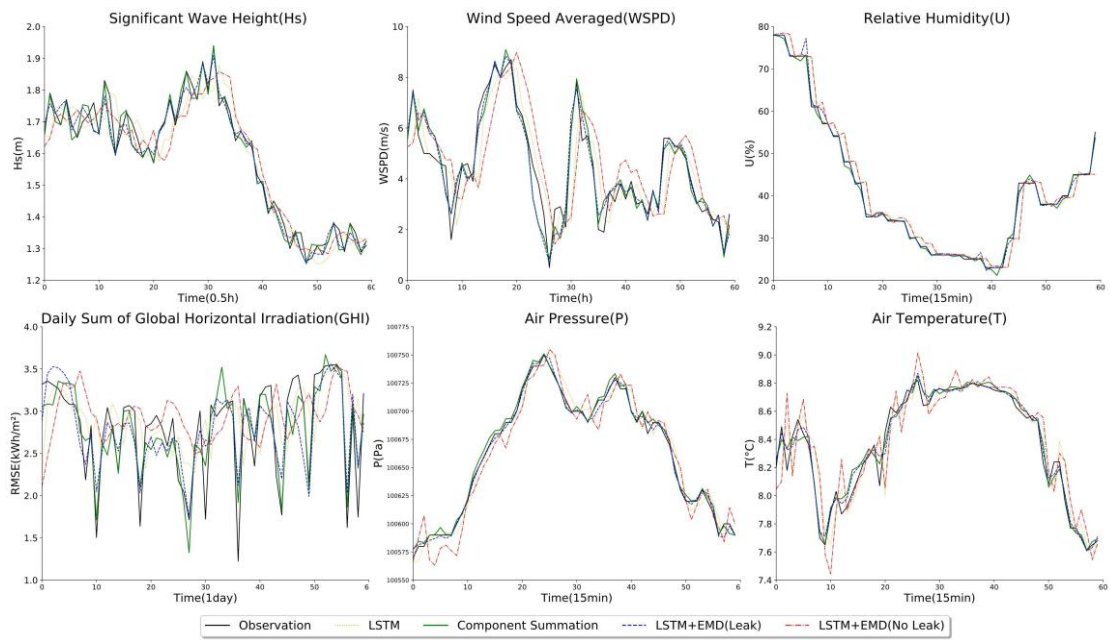

Figure E.19: The groundtruth and predicted curves of a randomly selected sample on different datasets. We compare the component summation model to the original sequence, the LSTM model, and the EMD model with and without leakage. The version with label leakage has much more similar shapes with the groundtruth curve than the version without leakage.

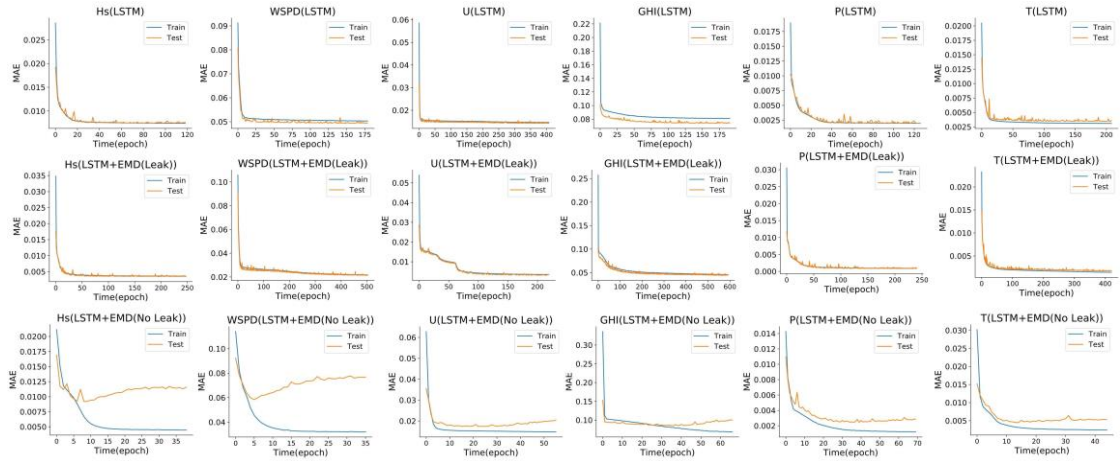

Figure E.20: The curves of training and test loss over different model training epochs. The test losses of the models with information leakage are substantially smaller than those without leakage. The test loss curves of the non-leaked version have evident inflection points that indicate the overfitting phenomenon, while they cannot be observed on the test loss curves of the leaked version.